\documentclass[runningheads]{llncs}
\usepackage{graphicx}
\usepackage{amsmath}
\usepackage{amsmath,bm}
\usepackage{booktabs}
\usepackage{multirow}
\usepackage{graphicx}
\usepackage{amssymb}
\usepackage{cite} 
\usepackage{bbding}
\usepackage{float} 
\usepackage{enumerate}
\usepackage{siunitx}
\begin{document}
\title{DTWSSE: Data Augmentation with a\\ Siamese Encoder for Time Series}
%
%
\author{Xinyu Yang \orcidID{0000-0003-0481-4880} \and Xinlan Zhang \and 
Zhenguo Zhang\thanks{Corresponding author} \and \\ Yahui Zhao \and Rongyi Cui}
\authorrunning{X. Yang et al.}
%
\institute{Department of Computer Science and Technology, Yanbian University \\
    977 Gongyuan Road, Yanji, 133002, People’s Republic of China \\
    \email{\{2020050049, 2019050390, zgzhang, yhzhao, cuirongyi\}@ybu.edu.cn}
}
 
\maketitle              
\begin{abstract}
Access to labeled time series data is often limited in the real world, which constrains the performance of deep learning models in the field of time series analysis. Data augmentation is an effective way to solve the problem of small sample size and imbalance in time series datasets. The two key factors of data augmentation are the distance metric and the choice of interpolation method. SMOTE does not perform well on time series data because it uses a Euclidean distance metric and interpolates directly on the object. Therefore, we propose a DTW-based synthetic minority oversampling technique using siamese encoder for interpolation named DTWSSE. In order to reasonably measure the distance of the time series, DTW, which has been verified to be an effective method forts, is employed as the distance metric. To adapt the DTW metric, we use an autoencoder trained in an unsupervised self-training manner for interpolation. The encoder is a Siamese Neural Network for mapping the time series data from the DTW hidden space to the Euclidean deep feature space, and the decoder is used to map the deep feature space back to the DTW hidden space. We validate the proposed methods on a number of different balanced or unbalanced time series datasets. Experimental results show that the proposed method can lead to better performance of the downstream deep learning model.

\keywords{Data augmentation \and Time series \and Siamese encoder \and Dynamic time warping.}
\end{abstract}

\section{Introduction}
Time series analysis is an important task in the field of data mining~\cite{fu2011review}. Deep learning is now widely used in time series analysis and has achieved significant success~\cite{gamboa2017deep, fawaz2019deep}. Since in many cases the cost of acquiring labeled time series data is large and the sample size of different classes of time series data is very imbalanced~\cite{wen2020time, jiang2019gan}, deep learning models are likely to overfit or ignore the minority classes during training. The most effective way to solve these problems is to perform data augmentation by oversampling synthetic samples. Due to the high dimensionality of time series~\cite{zhu2020oversampling}, the dimensions are time-correlated and it is difficult to compare the similarity between different time series~\cite{abid2018autowarp}, which makes data augmentation of time series data challenging.

SMOTE is one of the classical data augmentation methods~\cite{fernandez2018smote, chawla2002smote}. However, the Euclidean distance used in this method does not represent the similarity between time series well, and direct interpolation of two time series may impair the temporal correlation between dimensions within the data.

In this work, we propose a novel method called \textbf{DTW}-based \textbf{S}ynthetic minority oversampling technique using \textbf{S}iamese \textbf{E}ncoder for interpolation (DTWSSE). Based on the classical SMOTE method for possible phase shifts and amplitude changes of the time series, we use the DTW~\cite{rakthanmanon2012searching} metric to measure the distance between time series. The encoder and decoder are trained in an unsupervised self-training manner. The encoder will learn the mapping relationship from DTW hidden space to Euclidean deep feature space, and the decoder will learn the mapping relationship from Euclidean deep feature space back to DTW hidden space. The process of interpolation on the Euclidean deep feature space and decoder mapping to the DTW hidden space is adapted to the DTW metric and better preserves the temporal properties of the time series. We validate our proposed method and its components on a number of time series datasets from different domains. The experimental results show that the components of our proposed method coordinate with each other to produce significant performance improvements in downstream deep learning models.

\section{Related work}
Data augmentation aims to improve the performance of downstream models by synthesizing data~\cite{fawazdata}. SMOTE is a widely used data augmentation method for imbalanced data, which randomly selects some ``center" samples from the minority of classes and interpolates between the ``center" and its K-nearest neighbors to synthesize new data~\cite{fernandez2018smote,chawla2002smote}. This method is based on Euclidean distance and direct interpolation of samples, which is contradictory to time series characteristics when dealing with time series data. Smart augmentation is a way to augment the data by adding a generator before the downstream model, which will intelligently synthesize new samples for training based on the downstream model~\cite{lemley2017smart}. The generators of this method are trained with supervised learning and downstream models together. However, this approach causes higher costs when training downstream models. 

Dynamic time warping (DTW) is widely used for time series distance metric~\cite{rakthanmanon2012searching}, which is aimed at the possible phase shift and amplitude change of time series, and it uses dynamic programming to align data at different time points to achieve a reasonable comparison of time series similarity. Autowarp is an end-to-end approach to learning better metric~\cite{abid2018autowarp}. This method learns better metric from an unlabeled time series dataset by unsupervised learning. However, the model of this method needs to be pre-trained for each dataset. 

Several studies have shown that interpolation in the deep feature space is feasible. \textit{Upchur et al.}~\cite{upchurch2017deep} shows that interpolation in deep feature space enables image semantic changes. \textit{DeVries et al.}~\cite{devries2017dataset} shows that interpolation in feature space is superior to direct interpolation of objects. The feature spaces used by these methods are randomly generated by the model corresponding to the method.

The original design of Siamese neural network is to calculate the similarity of two inputs by mapping the two inputs to a new feature space through a neural network~\cite{chicco2021siamese}. When using Contrastive loss, the model will make objects of the same class as close as possible in the feature space and objects of different classes as far as possible. The method proposed by \textit{Utkin et al}. is to interpolate the output deep feature space of the siamese network~\cite{utkin2019explanation}. However, the autoencoder used in this approach requires supervised learning using labeled data and cannot be used in cases where the amount of data is small.

\section{The Proposed DTWSSE Method}
The general architecture of DTWSSE is based on the classical SMOTE technology. The core idea of DTWSSE is to use the DTW metric, which has been verified as a valid metric for time series distance. Moreover, DTWSSE uses an autoencoder that has undergone a special unsupervised training process in order to adapt the DTW metric for the interpolation operation.

\subsection{Make the dataset balanced}
\resizebox{\textwidth}{!}{
In this paper, each time series sample is denoted as $\bm{X_i}=\left \lbrack x^1,x^2,...,x^L \right \rbrack$
}\\
\noindent$\left( i=1,2...,N \right)$, where $x^i \in \mathbb{R}^M \left( i=1,2,...,L \right)$ and $N$ is the sample size of the dataset. That is, each sample is an ordered collection of M-dimensional values of the length $L$, thus $\bm{X_i} \in \mathbb{R}^{L \times M}$. Each sample in the dataset has a unique label $C^i \left( i=1,2,...,N \right)$.

Suppose there are $c$ classes in the dataset, and the dataset is needs to be augmented 
by a multiplier of $T$. We consider how to handle the data belonging to a class, firstly.
To ensure the balance of classes in the augmented dataset, if the sample size of the class $\mathcal A$
is $a$, then the number of samples to be added by oversampling for this class is $onum_{\mathcal A}$.

\begin{equation}
  onum_{\mathcal A}=\left \lfloor \frac{N \cdot T}{c} \right \rfloor-a 
\end{equation}

\noindent Next, we randomly select some ``centers" within the class and use the \textit{KNN} algorithm based on 
the DTW to obtain the $k$ nearest neighbors of each ``center". The number of ``centers" selected 
in the class $\mathcal A$ is shown below. Note that the process of finding nearest neighbors is 
only done within the class.

\begin{equation}
  cnum_{\mathcal A}=\left \lceil \frac{onum_{\mathcal A}}{k} \right \rceil
\end{equation}

\subsection{Use DTW to select instances for new data generation}
We let a ``center" be $\bm{X_q}=[x^1_q,x^2_q,...,x^L_q]$, and another sample belonging to the class
$\mathcal A$ is $\bm{X_s}=[x^1_s,x^2_s,...,x^L_s]$. To calculate the DTW distance between $\bm{X_q}$
and $\bm{X_s}$, we first construct an $L \times L$ matrix $D$, where $D_{ij}$ denotes the cost of 
aligning $X_q^i$ to $X_s^j$.
\begin{equation}
  D_{ij}= {\left \| x_q^i-x_s^j \right\|}_2^2
\end{equation}
The warping path $W=\left[ w_1,w_2,...,w_p,...,w_P \right] \left( L \le p < 2L \right)$ is the sequence
of grid points, where each $w_p= \left( i_p,j_p\right)$ corresponds to an element $D_{ij}$ of the matrix
$D$. In addition, any warping path needs to satisfy the following three constraints:
\begin{enumerate}[(1)]
  \sloppy
  \item Boundary conditions: $w_1= \left( 1,1\right)$, and $w_p= \left( L,L\right)$
  \item Monotonicity condition: $1=i_1 \le i_2 \le ... \le i_P=L$ and $1=j_1 \le j_2 \le ... \le j_P=L$
  \item Valid step: $w_p= \left( i_p,j_p \right)$  $\implies$ \\ $w_{p+1} \in  \left\{ \left( i_p+1,j_p \right) , \left( i_p,j_p+1 \right), \left( i_p+1,j_p+1 \right) \right\}$
\end{enumerate}

The warping distance $d$ is a function that maps a warping path to a non-negative real number.
\begin{equation}
  d \left( \bm{X_q},\bm{X_s} \right)= \sum_{t=1}^{P}D_{i_t j_t} 
\end{equation}
The DTW is the minimum value of the warping distance corresponding to all feasible warping paths.
\begin{equation}
  DTW\left( \bm{X_q},\bm{X_s} \right)=min_W \left\{ \sum_{t=1}^P D_{i_tj_t} \right\}
\end{equation}
We define $dp\left( i,j \right)$ as the cumulative distance when moving to the element $D_{ij}$ according to the warping path corresponding to DTW, so that the DTW can be calculated by the following dynamic programming formulation.
\begin{equation}
  dp\left( i,j \right)=D_{ij}+min \left\{ dp\left( i-1,j \right) , dp\left( i,j-1 \right) , dp\left( i-1,j-1 \right)  \right\} 
\end{equation}
After calculating the DTW of $\bm{X_q}$ with other samples in class $\mathcal A$,
the $k$ nearest neighboring samples with the DTW distance to $\bm{X_q}$ are obtained.
We assume that one of these $k$ nearest neighbors is $\bm{X_e}$.
Since we use the DTW to measure the distance between $\bm{X_q}$ and $\bm{X_e}$,
we need to oversampling using the appropriate interpolation method corresponding to the DTW metric.

\subsection{Interpolation adapted to the DTW metric}
In this work, we use an autoencoder to achieve oversampling by interpolating on the Euclidean 
deep feature space. The encoder is a siamese neural network designed to learn to map data from the DTW hidden space to the Euclidean feature space, so that the DTW metric between time series is equivalent to the Euclidean metric between latent vectors output by encoder. 
The decoder is a neural network that maps data from the Euclidean deep feature space back to the DTW hidden space.

We use a special unsupervised learning approach to train the autoencoder. The dataset for training the autoencoder is $\mathcal D=\left\{ \left[ \left( \bm{S_i^1},\bm{S_i^2} \right),y_i \right]  \right\}$, where $\bm{S_i^1}$ and $\bm{S_i^2}$ are randomly generated sequences and $\bm{S_i^1},\bm{S_i^2} \in \mathbb{R}^{L \times M}$. $y_i$ is a label that is automatically generated after generating $\bm{S_i^1}$ and $\bm{S_i^2}$, which represents the DTW between $\bm{S_i^1}$ and $\bm{S_i^2}$.

\begin{figure}
  \centering
  \vspace{-9mm}  
  \setlength{\belowcaptionskip}{-6mm}   
  \includegraphics[height=6cm]{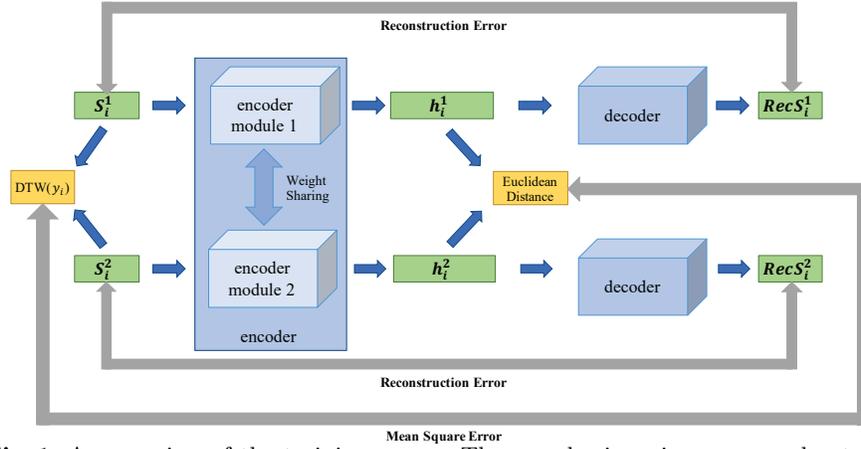}
  \vspace{-6mm}
  \caption{An overview of the training process.The encoder is a siamese neural network that inputs two vectors at a time and generates two latent vectors. The decoder inputs one latent vector at a time and produces a reconstructed input vector. When training the encoder, the goal is to minimize the mean square error of the DTW distance between the input vectors and the Euclidean distance between the output vectors. When training the decoder, the goal is to minimize the reconstruction error of autoencoder.}
  \label{fig:fig1}
\end{figure}

As shown in Fig.~\ref{fig:fig1}, the sequences $\bm{S_i^1}$ and $\bm{S_i^2}$ will be input to encoder and generate the corresponding latent vectors $\bm{h_i^1}$ and $\bm{h_i^2}$.
\vspace{-3mm}
\begin{equation}
  \bm{h_i^1},\bm{h_i^2}=encoder\left( \bm{S_i^1} , \bm{S_i^2} \right)
\end{equation}

When training the encoder, our goal is to minimize the mean square error between 
the Euclidean distance of $\bm{h_i^1}$ and $\bm{h_i^2}$ and the DTW of $\bm{S_i^1}$ and $\bm{S_i^2}$.
Therefore, the loss function when we train the encoder can be expressed in the following form. 
\vspace{-2mm}
\begin{equation}
  L_E=\frac{1}{\left| \mathcal D \right|} \sum_i \left( {\left\| \bm{h_i^1}-\bm{h_i^2} \right\|}_2 - y_i \right)^2
\end{equation}

\noindent It is important to emphasize that during the training process, the two modules of 
inside the encoder have the same structure and share weights, 
in other words, the encoder is a \textbf{Siamese neural network}. From another point of view, 
one of the encoder modules acts as a discriminator to the output of the other module.

The decoder will input two latent vectors $\bm{h_i^1}$ and $\bm{h_i^2}$, respectively, 
and output the reconstructed sequences $\bm{RecS_i^1}$ and $\bm{RecS_i^2}$.
\vspace{-2mm}
\begin{equation}
\begin{split}
    \bm{RecS_i^1}=decoder \left( \bm{h_i^1} \right) \\
    \bm{RecS_i^2}=decoder \left( \bm{h_i^2} \right)
\end{split}
\end{equation}
Our goal in training the decoder is to minimize the reconstruction error between the 
output of the decoder and the input of the encoder. Thus, the loss function when training 
the decoder can be expressed in the following form. It is important to note that when 
training the decoder we need to fix the parameters of the encoder.
\begin{equation}
  L_D=\frac{1}{2\left| \mathcal D \right|}\sum_i\left( 
    {\left\| \bm{S_i^1}-\bm{RecS_i^1} \right\|}_2 \right) ^2 +\frac{1}{2\left| \mathcal D \right|}
    \sum_i\left( {\left\| \bm{S_i^2}-\bm{RecS_i^2} \right\|}_2 \right) ^2
\end{equation}
In practice, we first train the encoder until $L_E$ converges, then train the 
decoder to learn to undo the mapping relationship. In addition, \textbf{we can use 
different architectures of deep neural networks as encoder module and decoder}, such as
CNN, Fully-connected Neural Network.

\begin{figure}
  \centering
   \vspace{-13mm}  
  \setlength{\belowcaptionskip}{-6mm}   
  \includegraphics[height=4cm]{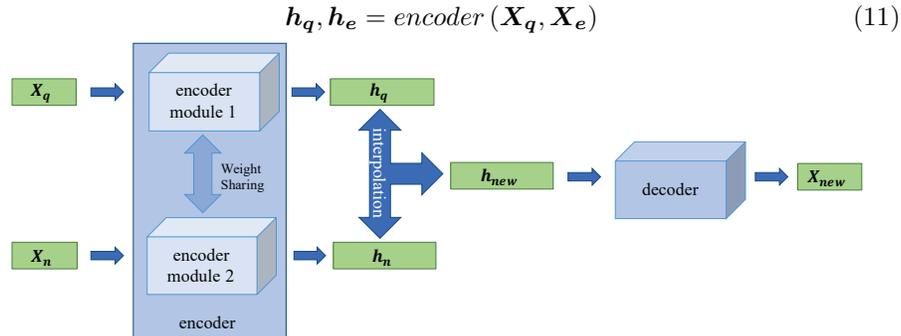}
  \vspace{-5mm}
  \caption{An overview of the interpolation process. Each time the encoder inputs two time series and generates the corresponding latent vectors, we interpolate the two latent vectors and the decoder decodes them to form the new time series synthetic data.}
  \label{fig:fig2}
\end{figure}

After training, we use the above autoencoder for oversampling. As shown in 
Fig.~\ref{fig:fig2}, when we input $\bm{X_q}$ and $\bm{X_e}$ to the encoder, two latent 
vectors $\bm{h_q}$ and $\bm{h_e}$ will be generated correspondingly.
\begin{equation}
  \bm{h_q},\bm{h_e}=encoder\left(\bm{X_q} , \bm{X_e} \right) 
\end{equation}

Since the latent vector is in the Euclidean deep feature space, 
it is reasonable to perform interpolation between the latent vectors. 
The new sample $\bm{X_{new}}$ generated by oversampling and its corresponding latent vector
$\bm{h_{new}}$ can be derived from the following equations.
\begin{equation}
    \bm{h_{new}}=\bm{h_q} + rand \left( 0,1 \right) \cdot \left( \bm{h_e}-\bm{h_q} \right) 
\end{equation}
\begin{equation}
  \bm{X_{new}}=decoder\left(\bm{h_{new}}\right)
\end{equation}
Since the decoder maps the latent vector from the Euclidean deep feature space back to the 
DTW hidden space, we implement an interpolation procedure that matches the DTW metric. 

After interpolating once between all the ``centers" selected in class $\mathcal A$ and 
their $k$ nearest neighbors, we have completed the data augmentation of class $\mathcal A$.
After the above process is done for all classes, we get the augmented time series dataset.

\section{Experiment}

\begin{figure} 
  \vspace{-9mm}
  \setlength{\belowcaptionskip}{-6mm}   
  \centering
  \includegraphics[width=\textwidth]{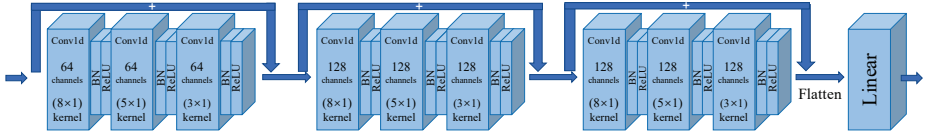}
  \vspace{-6mm}
  \caption{The schematic diagram of the ResNet classifier architecture.}
  \label{fig:fig3}
\end{figure}

In this section, we validate the proposed method on several datasets for classification 
tasks. For all datasets, we use the same ResNet classifier~\cite{wang2017time}, whose architecture is shown in Fig.~\ref{fig:fig3}. The only difference is that the data for training the classifier was obtained by different data augmentation methods.

In each experiment, to ensure the consistency of the experimental variables, the classifier is trained with 100 epochs using a mini-batch stochastic gradient descent method with a batch size of $32$ for each experiment. We trained the classifier using the \emph{Adam} optimizer with the hyper parameter \emph{lr} set to 1e-3, the loss function is the cross entropy between the output of the classifier and the corresponding label. In our previous work we found that larger expansion multipliers are more likely to yield good results, so when preprocessing  dataset we set the expansion multiplier $T$ to $10$. In addition, we only consider the closest sample to the ``center" in our experiments.

\begin{figure} 
  \centering
  \vspace{-0.8cm}  
  \setlength{\belowcaptionskip}{0mm}   
  \includegraphics[width=\textwidth]{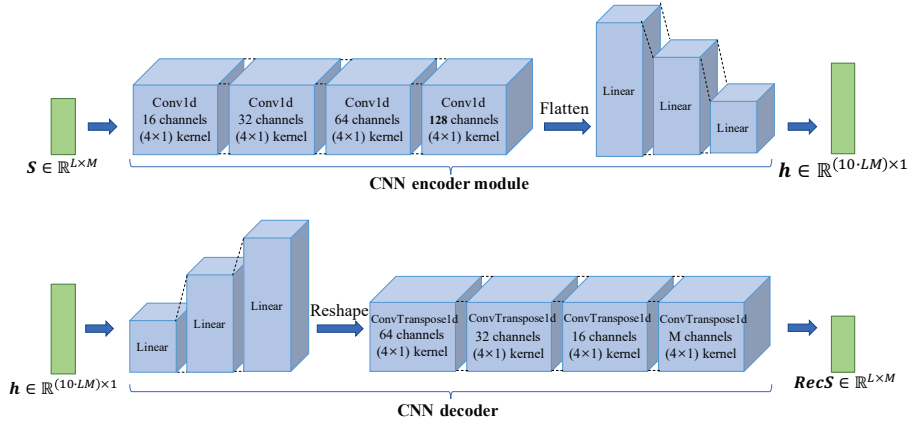}
  \vspace{-8mm}
  \caption{The schematic diagram of autoencoder for CNN architecture. There are two encoder modules with the same structure and shared weights in one encoder. Decoder architecture is completely symmetric with encoder module.}
  \label{fig:fig4}
\end{figure}

\begin{figure} 
  \centering
  \vspace{-0.8cm}  
  \setlength{\belowcaptionskip}{0mm}   
  \includegraphics[width=\textwidth]{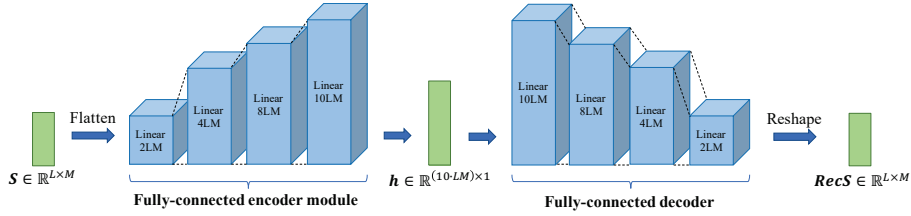}
  \vspace{-8mm}
  \caption{The schematic diagram of autoencoder for Fully-connected Neural Network architecture. There are two encoder modules with the same structure and shared weights in one encoder. Decoder architecture is completely symmetric with encoder module.}
  \label{fig:fig5}
\end{figure}

The specific architecture of the autoencoder used in the experiments is related to the 
dimension of the time series in the dataset. In some previous work, we found that 
the convergence of the loss function of autoencoder in the proposed method is easier when the dimension of the latent vector is 10 times the number of variables of the time series.
So if the time series data $\bm{X_i} \in \mathbb{R}^{L \times M}$, we set the dimension of the latent vector to $10\cdot L\cdot M$, i.e. $\bm{h_i} \in \mathbb R^{\left( 10\cdot L\cdot M \right) \times 1}$.
In addition, encoder can be replaced with different architectures of neural networks, while 
the architecture of decoder should be symmetric with encoder. 
In our experiments we used CNN and Fully-connected Neural Network respectively for comparison, their architectures are shown in Fig.~\ref{fig:fig4} and Fig.~\ref{fig:fig5}.

We performed each independent experiment 10 times and recorded 3 assessment metrics, 
the classifier was trained from scratch for each experiment. The \textbf{Top-1 accuracy} represents
the best result the model can obtain, the \textbf{worst accuracy} represents the worst result the model can obtain, and the \textbf{average accuracy} with the best and worst accuracy removed represents the result the model can obtain in most cases. All results are obtained with a single NVIDIA$^\circledR$ Tesla$^\circledR$ V100 GPU.

\vspace{-4mm}
\subsection{Basic results}
\vspace{-2mm}

\begin{table} 
\vspace{-8mm}
\caption{Summary of the time series datasets used in the experiments. The ``\#" in the table is short for ``quantity".}
\label{tab1}
\centering
\vspace{-3mm}
\resizebox{\textwidth}{!}{
\begin{tabular}{@{}lccccc@{}}
\toprule
\textbf{Datasets}              & \textbf{\begin{tabular}[c]{@{}c@{}}Train\\ Size\end{tabular}} & \textbf{\begin{tabular}[c]{@{}c@{}}Test\\ Size\end{tabular}} & \textbf{\begin{tabular}[c]{@{}c@{}}\# of\\ Classes\end{tabular}} & \textbf{\begin{tabular}[c]{@{}c@{}}\# of\\ Minority Class\end{tabular}} & \textbf{\begin{tabular}[c]{@{}c@{}}\# of\\ Majority Class\end{tabular}} \\ \midrule
DistalPhalanxTW                & 400                                                           & 139                                                          & 6                                                                & 28                                                                      & 82                                                                      \\
DistalPhalanxOutlineAgeGroup   & 400                                                           & 139                                                          & 3                                                                & 30                                                                      & 257                                                                     \\
DistalPhalanxOutlineCorrect    & 600                                                           & 276                                                          & 2                                                                & 222                                                                     & 378                                                                     \\
MiddlePhalanxTW                & 399                                                           & 154                                                          & 6                                                                & 30                                                                      & 84                                                                      \\
MiddlePhalanxOutlineAgeGroup   & 400                                                           & 154                                                          & 3                                                                & 55                                                                      & 237                                                                     \\
MiddlePhalanxOutlineCorrect    & 600                                                           & 291                                                          & 2                                                                & 212                                                                     & 338                                                                     \\
ProximalPhalanxTW              & 400                                                           & 205                                                          & 6                                                                & 16                                                                      & 109                                                                     \\
ProximalPhalanxOutlineAgeGroup & 400                                                           & 205                                                          & 3                                                                & 72                                                                      & 189                                                                     \\
ProximalPhalanxOutlineCorrect  & 600                                                           & 291                                                          & 2                                                                & 194                                                                     & 406                                                                     \\
PhalangesOutlinesCorrect       & 1800                                                          & 858                                                          & 2                                                                & 628                                                                     & 1172                                                                    \\ \bottomrule
\end{tabular}
}
\vspace{-6mm}
\end{table}

We used 10 time series datasets to conduct experiments comparing the proposed method with the classical SMOTE method. These datasets are derived from images of hand bones and extracted by \textit{Cao et al.}~\cite{davis2013predictive}. The feature numbers of the data in the above datasets are all 80, and the details are shown in Table~\ref{tab1}. It is clear that these datasets are imbalanced. 

In our experiments, we compared the results generated by the classifier trained on the 
original dataset, the dataset augmented by the classical SMOTE method, and the dataset 
augmented by the DTWSSE, respectively. We also documented the results of using autoencoder 
with different architectures. Since the dimension of all the data in the above datasets is $80 \times 1$, we set the dimension of the latent vector generated by encoder to $800 \times 1$.
The experimental results are shown in Table~\ref{tab2}. 

\begin{table}
\vspace{-6mm}
\caption{The performance results obtained by training the classifier with the dataset augmented by different methods. The ``original" column indicates the result of training the downstream classifier using the original dataset without any data augmentation methods.} 
\label{tab2}
\vspace{-3mm}
\centering
\resizebox{!}{6.5cm}{
\begin{tabular}{@{}c|l|cc|cc@{}}
\toprule
\multicolumn{1}{l|}{\textbf{Metrics}}                                        & \textbf{Datasets} & \textbf{Original} & \textbf{SMOTE} & \multicolumn{1}{c}{\textbf{\begin{tabular}[c]{@{}c@{}}DTWSSE\\ (CNN)\end{tabular}}} & \multicolumn{1}{c}{\textbf{\begin{tabular}[c]{@{}c@{}}DTWSSE\\ (Fully-connected)\end{tabular}}} \\ \midrule
\multirow{10}{*}{\begin{tabular}[c]{@{}c@{}}Top-1\\ Accuracy(\%)\end{tabular}}   & Dis.TW            & 64.75             & 69.06          & 69.06                & \textbf{74.10}       \\
                                                                             & Dis.A.G.          & 74.82             & 75.54          & \textbf{76.98}       & 76.26                \\
                                                                             & Dis.C.            & 79.71             & 79.71          & \textbf{80.07}       & 77.90                \\
                                                                             & Mid.TW            & 53.90             & 52.60          & 56.49                & \textbf{57.14}       \\
                                                                             & Mid.A.G.          & \textbf{64.94}    & 56.49          & 62.34                & 62.99       \\
                                                                             & Mid.C.            & \textbf{84.19}    & 83.85          & \textbf{84.19}       & 83.16                \\
                                                                             & Pro.TW            & \textbf{80.98}    & 79.02          & \textbf{80.98}       & 80.49                \\
                                                                             & Pro.A.G.          & 87.32             & 87.80          & \textbf{88.29}       & 87.32                \\
                                                                             & Pro.C.            & 91.41             & 91.07          & \textbf{91.75}       & \textbf{91.75}       \\
                                                                             & Pha.              & 83.10             & 83.45 & \textbf{83.68}       & 82.98                \\ \midrule
\multirow{10}{*}{\begin{tabular}[c]{@{}c@{}}Worst\\ Accuracy(\%)\end{tabular}}   & Dis.TW            & 51.80    & \textbf{63.31} & 60.43                & \textbf{63.31}       \\
                                                                             & Dis.A.G.          & 47.48    & \textbf{67.63} & \textbf{67.63}       & 66.91                \\
                                                                             & Dis.C.            & 63.41             & 72.83          & \textbf{73.55}       & 71.01                \\
                                                                             & Mid.TW            & 28.57             & 45.45          & 46.75                & \textbf{50.00}       \\
                                                                             & Mid.A.G.          & 42.86             & 46.10          & \textbf{50.65}       & 46.10                \\
                                                                             & Mid.C.            & 52.58             & 59.11          & 74.57                & \textbf{74.91}       \\
                                                                             & Pro.TW            & 48.29             & 74.15          & \textbf{74.63}       & 72.68                \\
                                                                             & Pro.A.G.          & 73.66             & 80.98          & \textbf{82.44}       & 80.00                \\
                                                                             & Pro.C.            & 68.73             & 81.10          & 82.82                & \textbf{85.22}       \\
                                                                             & Pha.              & 55.13             & \textbf{79.60} & \textbf{79.60}       & 78.67                \\ \midrule
\multirow{10}{*}{\begin{tabular}[c]{@{}c@{}}Average\\ Accuracy(\%)\end{tabular}} & Dis.TW            & 59.44             & 65.20          & 66.19                & \textbf{67.72}       \\
                                                                             & Dis.A.G.          & 66.64             & 71.04          & \textbf{73.29}       & 72.57                \\
                                                                             & Dis.C.            & 72.51             & 76.18          & \textbf{76.95}       & 75.27                \\
                                                                             & Mid.TW            & 46.35             & 49.11          & 52.60                & \textbf{52.84}       \\
                                                                             & Mid.A.G.          & 54.46             & 51.22          & \textbf{57.63}       & 56.25                \\
                                                                             & Mid.C.            & 77.41             & 79.38          & \textbf{79.98}       & 79.73                \\
                                                                             & Pro.TW            & 77.38             & 76.89          & \textbf{78.90}       & 76.95                \\
                                                                             & Pro.A.G.          & 83.78    & \textbf{85.55} & 85.06                & 85.18                \\
                                                                             & Pro.C.            & 83.93             & 87.41 & \textbf{88.92}       & 87.46                \\
                                                                             & Pha.              & 73.86             & 81.15 & \textbf{81.22}       & 80.36                \\ \bottomrule
\end{tabular}
}
\vspace{-8mm}
\end{table}

It is observed that our proposed data augmentation method DTWSSE can significantly improve the classifier performance in most cases. Compared with the classical SMOTE method, DTWSSE leads to a significant improvement in the performance of the classifier. For example, on the \textit{DistalPhalanxTW} dataset, the DTWSSE method using the DNN architecture autoencoder improved the Top-1 accuracy by 9.35\% (74.10\% v.s. 64.75\%), the worst accuracy by 11.51\% (63.31\% v.s. 51.80\%), and the average accuracy by 8.28\% (67.72\% v.s. 59.44\%), respectively, compared to the case of no data augmentation. Compared with the classical SMOTE method, the Top-1 accuracy is improved by 5.04\% (74.10\% v.s. 69.06\%), and the average accuracy is improved by 2.52\% (67.72\% v.s. 65.20\%). Although the worst accuracy is equal between the two methods (63.31\% v.s. 63.31\%), DTWSSE is more likely to produce better results.

Another thing that can be observed is that the optimal autoencoder architecture is different for different datasets. While our proposed approach leads to a general improvement in data augmentation, switching to a suitable autoencoder architecture can lead to a further improvement. This also inspired us to use the deep learning model that best fits the characteristics of the time series data in a practical task.

\vspace{-4mm}
\subsection{Ablation studies}
\vspace{-2mm}

\begin{table}  
\vspace{-8mm}
\caption{The results of the comparison with the SMOTE method using the DTW metric as the distance measure, the ``SMOTE(DTW)" column represents this method, which does not use autoencoder for the interpolation process.}
\label{tab3}
\vspace{-3mm}
\centering
\resizebox{!}{6.5cm}{
\begin{tabular}{@{}c|l|cc|cc@{}}
\toprule
\multicolumn{1}{l|}{\textbf{Metrics}}                                        & \textbf{Datasets} & \textbf{SMOTE} & \multicolumn{1}{c|}{\textbf{\begin{tabular}[c]{@{}c@{}}SMOTE\\ (DTW)\end{tabular}}} & \multicolumn{1}{c}{\textbf{\begin{tabular}[c]{@{}c@{}}DTWSSE\\ (CNN)\end{tabular}}} & \multicolumn{1}{c}{\textbf{\begin{tabular}[c]{@{}c@{}}DTWSSE\\ (Fully-connected)\end{tabular}}} \\ \midrule
\multirow{10}{*}{\begin{tabular}[c]{@{}c@{}}Top-1\\ Accuracy(\%)\end{tabular}}   & Dis.TW            & 69.06          & 69.06                                                                               & 69.06                & \textbf{74.10}       \\
                                                                             & Dis.A.G.          & 75.54          & 76.26                                                                               & \textbf{76.98}       & 76.26                \\
                                                                             & Dis.C.            & 79.71          & 78.26                                                                               & \textbf{80.07}       & 77.90                \\
                                                                             & Mid.TW            & 52.60          & 51.30                                                                               & 56.49                & \textbf{57.14}       \\
                                                                             & Mid.A.G.          & 56.49          & 54.55                                                                               & 62.34                & \textbf{62.99}       \\
                                                                             & Mid.C.            & 83.85          & 81.44                                                                               & \textbf{84.19}       & 83.16                \\
                                                                             & Pro.TW            & 79.02          & 80.49                                                                               & \textbf{80.98}       & 80.49                \\
                                                                             & Pro.A.G.          & 87.80          & 86.34                                                                               & \textbf{88.29}       & 87.32                \\
                                                                             & Pro.C.            & 91.07          & 91.07                                                                               & \textbf{91.75}       & \textbf{91.75}       \\
                                                                             & Pha.              & 83.45          & \textbf{83.92}                                                                      & 83.68                & 82.98                \\ \midrule
\multirow{10}{*}{\begin{tabular}[c]{@{}c@{}}Worst\\ Accuracy(\%)\end{tabular}}   & Dis.TW            & \textbf{63.31} & 58.99                                                                               & 60.43                & \textbf{63.31}       \\
                                                                             & Dis.A.G.          & \textbf{67.63} & 62.59                                                                               & \textbf{67.63}       & 66.91                \\
                                                                             & Dis.C.            & 72.83          & 73.19                                                                               & \textbf{73.55}       & 71.01                \\
                                                                             & Mid.TW            & 45.45          & 46.75                                                                               & 46.75                & \textbf{50.00}       \\
                                                                             & Mid.A.G.          & 46.10          & 44.81                                                                               & \textbf{50.65}       & 46.10                \\
                                                                             & Mid.C.            & 59.11          & 58.76                                                                               & 74.57                & \textbf{74.91}       \\
                                                                             & Pro.TW            & 74.15          & 69.27                                                                               & \textbf{74.63}       & 72.68                \\
                                                                             & Pro.A.G.          & 80.98          & 80.98                                                                               & \textbf{82.44}       & 80.00                \\
                                                                             & Pro.C.            & 81.10          & 79.38                                                                               & 82.82                & \textbf{85.22}       \\
                                                                             & Pha.              & 79.60          & \textbf{80.77}                                                                      & 79.60                & 78.67                \\ \midrule
\multirow{10}{*}{\begin{tabular}[c]{@{}c@{}}Average\\ Accuracy(\%)\end{tabular}} & Dis.TW            & 65.20          & 63.76                                                                               & 66.19                & \textbf{67.72}       \\
                                                                             & Dis.A.G.          & 71.04          & 71.13                                                                               & \textbf{73.29}       & 72.57                \\
                                                                             & Dis.C.            & 76.18          & 75.77                                                                               & \textbf{76.95}       & 75.27                \\
                                                                             & Mid.TW            & 49.11          & 49.19                                                                               & 52.60                & \textbf{52.84}       \\
                                                                             & Mid.A.G.          & 51.22          & 49.27                                                                               & \textbf{57.63}       & 56.25                \\
                                                                             & Mid.C.            & 79.38          & 77.71                                                                               & \textbf{79.98}       & 79.73                \\
                                                                             & Pro.TW            & 76.89          & 76.95                                                                               & \textbf{78.90}       & 76.95                \\
                                                                             & Pro.A.G.          & \textbf{85.55} & 84.63                                                                               & 85.06                & 85.18                \\
                                                                             & Pro.C.            & 87.41          & \textbf{89.69}                                                                      & 88.92                & 87.46                \\
                                                                             & Pha.              & 81.15          & \textbf{81.95}                                                                      & 81.22                & 80.36                \\ \bottomrule
\end{tabular}
}
\vspace{-8mm}
\end{table}

In order to better understand the effectiveness of each component of DTWSSE, we conducted the following two ablation studies. We still used the above 10 datasets for our experiments.

First we considered the case of using the DTW metric and interpolating directly between time series. The experimental results are shown in the Table~\ref{tab3}. We can clearly observe that most of the experimental results are degraded compared to DTWSSE because the autoencoder interpolation method adapted to DTW metric is not used. Compared to the classical SMOTE method, we cannot clearly distinguish which of the two methods is superior. For Top-1 Accuracy, the classical SMOTE method wins 5 times and the SMOTE method using the DTW metric wins 3 times; for Worst Accuracy, the ratio is 6:3; and for Average Accuracy, the ratio is 5:5. This may indicate that after finding the nearest neighbor of the ``center" using the DTW distance, interpolation between the nearest neighbor and the ``center" on the DTW feature space is necessary.

\begin{figure}
  \centering
  \vspace{-8mm}  
  \setlength{\belowcaptionskip}{-6mm}   
  \includegraphics[height=2cm]{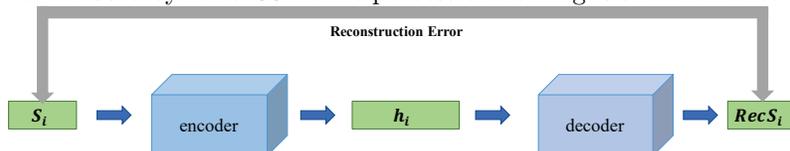}
  \vspace{-4mm}
  \caption{An overview of the naive autoencoder training process. This approach does not use the siamese network as the encoder architecture, and the encoder does not have to learn any special mapping relations. In this ablation study, we only minimize the reconstruction error of autoencoder.}
  \label{fig:fig6}
\end{figure}

We also considered the case of using naive autoencoder to replace the autoencoder used in DTWSSE. The only difference with DTWSSE is the unsupervised training process of the autoencoder. For this experiment, we did not train encoder to learn how to map the DTW hidden space to the Euclidean deep feature space. Instead of designing the encoder as a siamese network and fixing the parameters of a certain part, we trained the encoder and decoder together to minimize the reconstruction error. At this point, encoder is actually a module of the encoder used by DTWSSE. The process of training is shown in the Fig.~\ref{fig:fig6}.

The encoder and decoder still used the two architectures shown in Fig.~\ref{fig:fig4} and Fig.~\ref{fig:fig5}, but the encoder at this point is actually an encoder module in the figure. In the experiment, the dimension of hidden vector is still set to $800 \times 1$. 

The results are shown in the Table~\ref{tab4}. We find that this method outperforms the classical SMOTE method in most cases. This phenomenon suggests that interpolation on the deep feature space generated through the encoder can provide some improvement in performance. Compared to our proposed DTWSSE method, this method generally performs slightly worse. We note that when using autoencoder with CNN architecture, the method yields a worst accuracy of 34.42\% for the classifier after processing the \textit{MiddlePhalanxOutlineAgeGroup} dataset and a worst accuracy of 36.59\% for the classifier after processing the \textit{ProximalPhalanxOutlineAgeGroup} dataset. This suggests that interpolating only on the deep feature space may cause the model to converge at a large local minimum of the loss function during training. In addition our proposed method DTWSSE method performs better compared to this method in most cases. This phenomenon confirms the effectiveness of the component of the DTWSSE method that trains the encoder to learn the mapping relations from the DTW hidden space to the Euclidean deep feature space.

\begin{table}  
\vspace{-6mm}
\caption{The comparative results of the SMOTE method using the DTW metric as a distance measure and interpolation using a naive autoencoder. The column ``SMOTE+AE'' represents this approach, which uses autoencoder with CNN architecture and Fully-connected Neural Network architecture for comparison with DTWSSE, respectively.}
\label{tab4}
\centering
\vspace{-3mm}
\resizebox{\textwidth}{!}{
\begin{tabular}{@{}c|l|c|cc|cc@{}}
\toprule
\multicolumn{1}{l|}{\textbf{Metrics}}                                            & \textbf{Datasets} & \textbf{SMOTE} & \textbf{\begin{tabular}[c]{@{}c@{}}SMOTE+AE\\ (CNN)\end{tabular}} & \textbf{\begin{tabular}[c]{@{}c@{}}DTWSSE\\ (CNN)\end{tabular}} & \textbf{\begin{tabular}[c]{@{}c@{}}SMOTE+AE\\ (Fully-connected)\end{tabular}} & \textbf{\begin{tabular}[c]{@{}c@{}}DTWSSE\\ (Fully-connected)\end{tabular}} \\ \midrule
\multirow{10}{*}{\begin{tabular}[c]{@{}c@{}}Top-1 \\ Accuracy(\%)\end{tabular}}  & Dis.TW            & 64.75          & \textbf{69.06}                                                    & \textbf{69.06}                                                  & 70.50                                                             & \textbf{74.10}                                                  \\
                                                                                 & Dis.A.G.          & 74.82          & \textbf{76.98}                                                    & \textbf{76.98}                                                  & 74.10                                                    & \textbf{76.26}                                                  \\
                                                                                 & Dis.C.            & 79.71          & 79.35                                                             & \textbf{80.07}                                                  & 77.54                                                    & \textbf{77.90}                                                  \\
                                                                                 & Mid.TW            & 53.90          & 55.19                                                             & \textbf{56.49}                                                  & 56.49                                                             & \textbf{57.14}                                                  \\
                                                                                 & Mid.A.G.          & \textbf{64.94} & 59.09                                                             & \textbf{62.34}                                                          & \textbf{64.94}                                                    & 62.99                                                  \\
                                                                                 & Mid.C.            & 84.19          & \textbf{84.54}                                                    & 84.19                                                           & 82.13                                                    & \textbf{83.16}                                                  \\
                                                                                 & Pro.TW            & 80.98          & 80.49                                                             & \textbf{80.98}                                                  & \textbf{83.41}                                                    & 80.49                                                           \\
                                                                                 & Pro.A.G.          & 87.32          & \textbf{88.29}                                                    & \textbf{88.29}                                                  & \textbf{87.32}                                                    & \textbf{87.32}                                                  \\
                                                                                 & Pro.C.            & 91.41          & 91.41                                                             & \textbf{91.75}                                                  & 91.07                                                    & \textbf{91.75}                                                  \\
                                                                                 & Pha.              & 83.10          & 82.63                                                             & \textbf{83.68}                                                  & 82.28                                                             & \textbf{82.98}                                                  \\ \midrule
\multirow{10}{*}{\begin{tabular}[c]{@{}c@{}}Worst \\ Accuracy(\%)\end{tabular}}  & Dis.TW            & 51.80          & 51.80                                                    & \textbf{60.43}                                                  & \textbf{63.31}                                                    & \textbf{63.31}                                                  \\
                                                                                 & Dis.A.G.          & 47.48          & 66.19                                                    & \textbf{67.63}                                                  & \textbf{71.94}                                                    & 66.91                                                           \\
                                                                                 & Dis.C.            & 63.41          & 72.83                                                             & \textbf{73.55}                                                  & \textbf{72.10}                                                    & 71.01                                                           \\
                                                                                 & Mid.TW            & 28.57          & \textbf{46.75}                                                    & \textbf{46.75}                                                  & 48.05                                                             & \textbf{50.00}                                                  \\
                                                                                 & Mid.A.G.          & 42.86          & 34.42                                                             & \textbf{50.65}                                                  & 44.81                                                    & \textbf{46.10}                                                  \\
                                                                                 & Mid.C.            & 52.58          & 64.26                                                             & \textbf{74.57}                                                  & 62.20                                                             & \textbf{74.91}                                                  \\
                                                                                 & Pro.TW            & 48.29          & 71.71                                                             & \textbf{74.63}                                                  & 71.71                                                    & \textbf{72.68}                                                  \\
                                                                                 & Pro.A.G.          & 73.66          & 36.59                                                             & \textbf{82.44}                                                  & 79.02                                                    & \textbf{80.00}                                                  \\
                                                                                 & Pro.C.            & 68.73          & 81.44                                                             & \textbf{82.82}                                                  & 79.04                                                             & \textbf{85.22}                                                  \\
                                                                                 & Pha.              & 55.13          & 77.97                                                             & \textbf{79.60}                                                  & 77.74                                                             & \textbf{78.67}                                                  \\ \midrule
\multirow{10}{*}{\begin{tabular}[c]{@{}c@{}}Average\\ Accuracy(\%)\end{tabular}} & Dis.TW            & 59.44          & 64.57                                                             & \textbf{66.19}                                                  & 66.46                                                             & \textbf{67.72}                                                  \\
                                                                                 & Dis.A.G.          & 66.64          & 72.75                                                             & \textbf{73.29}                                                  & 72.48                                                    & \textbf{72.57}                                                  \\
                                                                                 & Dis.C.            & 72.51          & \textbf{76.95}                                                    & \textbf{76.95}                                                  & \textbf{75.72}                                                    & 75.27                                                           \\
                                                                                 & Mid.TW            & 46.35          & 49.68                                                             & \textbf{52.60}                                                  & 52.11                                                             & \textbf{52.84}                                                  \\
                                                                                 & Mid.A.G.          & 54.46          & 48.13                                                             & \textbf{57.63}                                                  & 56.09                                                    & \textbf{56.25}                                                  \\
                                                                                 & Mid.C.            & 77.41          & 79.47                                                             & \textbf{79.98}                                                  & 76.85                                                    & \textbf{79.73}                                                  \\
                                                                                 & Pro.TW            & 77.38          & 76.83                                                             & \textbf{78.90}                                                  & \textbf{77.74}                                                    & 76.95                                                           \\
                                                                                 & Pro.A.G.          & 83.78          & 82.62                                                    & \textbf{85.06}                                                  & 83.72                                                             & \textbf{85.18}                                                  \\
                                                                                 & Pro.C.            & 83.93          & \textbf{89.48}                                                    & 88.92                                                  & 86.64                                                             & \textbf{87.46}                                                  \\
                                                                                 & Pha.              & 73.86          & \textbf{81.47}                                                    & 81.22                                                  & \textbf{81.19}                                                    & 80.36                                                           \\ \bottomrule
\end{tabular}
}
\vspace{-8mm}
\end{table}

\vspace{-4mm}
\subsection{Apply to balanced datasets}
\vspace{-2mm}

We first conducted experiments on two traffic time series datasets, \textit{Chinatown} and \textit{MelbournePedestrian}\footnote{These datasets are available at http://www.timeseriesclassification.com/}. These datasets were recorded by automated pedestrian counting sensors located at various locations throughout the city of Melbourne, Australia. \textit{Hoang Anh Dau} edited the system-generated data over 12 months of 2017 to create these two datasets. Each sample has 24 features, representing the variation in the number of pedestrians in a day. The number on each dimension in sample represents the number of people captured by the sensor in one hour. In addition these datasets have been pre-segmented into training and test sets.

The samples in \textit{Chinatown} were recorded by the sensor at Chinatown-Swanston St. The dataset is divided into two classes, one from weekday and the other from weekend. The training set is balanced, with 10 samples for each of class, this is a typical few-shot learning problem. The test set size is 345.

The samples in \textit{MelbournePedestrian} were recorded by sensors located in different locations. Each class represents a location, and there are 10 classes in the dataset. The amount of samples for each class in the training set is about 120, and there are 1194 samples in total. There are 2439 samples in the test set.

With the expansion multiplier $T$ of 10, we conducted two comparison experiments and the results are shown in Table~\ref{tab5} and Table~\ref{tab6}. It can be observed that our proposed method still performs significantly better than the SMOTE on both datasets. In addition, for the \textit{Chinatown} dataset with a very small training sample size, the worst accuracy of the classifier is significantly lower than that without data augmentation as long as the data augmentation method is used. This may be because the synthetic samples make it more difficult for the downstream classifier to learn the true data distribution. However, since DTWSSE performs better on both the Top-1 accuracy metric and the average accuracy metric, this suggests that DTWSSE reduces the impact of this problem.
\begin{table} 
\vspace{-6mm}
\caption{The Performance results on the \textit{Chinatown} dataset.}
\label{tab5}
\centering
\vspace{-3mm}
\begin{tabular}{@{}cccc@{}}
\toprule
\textbf{Methods}     & \multicolumn{1}{c}{\textbf{\begin{tabular}[c]{@{}c@{}}Top-1\\ Accuracy(\%)\end{tabular}}} & \multicolumn{1}{c}{\textbf{\begin{tabular}[c]{@{}c@{}}Worst\\ Accuracy(\%)\end{tabular}}} & \multicolumn{1}{c}{\textbf{\begin{tabular}[c]{@{}c@{}}Average\\ Accuracy(\%)\end{tabular}}} \\ \midrule
No data augmentation & 98.54                                                                                     & \textbf{97.38}                                                                            & 98.03                                                                                       \\
SMOTE                & 98.83                                                                                     & 74.64                                                                                     & 95.85                                                                                       \\ \midrule
DTWSSE(CNN)          & \textbf{99.13}                                                                            & 89.21                                                                                     & \textbf{98.25}                                                                              \\
DTWSSE(DNN)          & \textbf{99.13}                                                                            & 84.84                                                                                     & 98.07                                                                                       \\ \bottomrule
\end{tabular}
\vspace{-10mm}
\end{table}

\begin{table} 
\caption{The Performance results on the \textit{MelbournePedestrian} dataset.}
\label{tab6}
\centering
\vspace{-3mm}
\begin{tabular}{@{}cccc@{}}
\toprule
\textbf{Methods}     & \multicolumn{1}{c}{\textbf{\begin{tabular}[c]{@{}c@{}}Top-1\\ Accuracy(\%)\end{tabular}}} & \multicolumn{1}{c}{\textbf{\begin{tabular}[c]{@{}c@{}}Worst\\ Accuracy(\%)\end{tabular}}} & \multicolumn{1}{c}{\textbf{\begin{tabular}[c]{@{}c@{}}Average\\ Accuracy(\%)\end{tabular}}} \\ \midrule
No data augmentation & 96.27                                                                                     & 87.25                                                                                     & 95.34                                                                                       \\
SMOTE                & 96.76                                                                                     & 84.26                                                                                     & 92.45                                                                                       \\ \midrule
DTWSSE(CNN)          & \textbf{97.13}                                                                            & 91.72                                                                                     & \textbf{96.26}                                                                              \\
DTWSSE(DNN)          & 96.72                                                                                     & \textbf{95.61}                                                                            & 96.20                                                                                       \\ \bottomrule
\end{tabular}
\vspace{-8mm}
\end{table}

We then conducted an experiment on the \textit{Libras} sign language dataset. \textit{Libras} Sign Language Movement Dataset is a dataset in the UC Irvine Machine Learning Repository~\cite{Dua:2019}, which is a multivariate dataset. There are 15 classes in the dataset, each class represents a hand movement type. There are 24 samples in each class and the total number of samples is 360. We randomly selected half of the samples from each class to form the training set, the rest of the samples to form the test set. Each sample is obtained from one video of the hand movement curve. During each recorded hand movement curves, 45 frames are selected and a time series sample is formed based on the two-dimensional coordinates of the hand center. In addition, each sample was subjected to time normalization in the unitary space. The experimental results are shown in the Table~\ref{tab7}. The results of this experiment also show that DTWSSE has better performance compared to SMOTE. These experiments also demonstrate the effectiveness of DTWSSE on balanced datasets.

\begin{table}
\vspace{-5mm}
\caption{The Performance results on the \textit{Libras} dataset.}
\label{tab7}
\centering
\vspace{-3mm}
\begin{tabular}{@{}cccc@{}}
\toprule
\textbf{Methods}     & \textbf{\begin{tabular}[c]{@{}c@{}}Top-1\\ Accuracy(\%)\end{tabular}} & \textbf{\begin{tabular}[c]{@{}c@{}}Worst\\ Accuracy(\%)\end{tabular}} & \textbf{\begin{tabular}[c]{@{}c@{}}Average\\ Accuracy(\%)\end{tabular}} \\ \midrule
No data augmentation & 95.56                                                                 & 85.00                                                                 & 92.36                                                                   \\
SMOTE                & 94.44                                                                 & 87.22                                                                 & 91.67                                                                   \\ \midrule
DTWSSE(CNN)          & \textbf{96.67}                                                        & 85.56                                                                 & 93.33                                                                   \\
DTWSSE(DNN)          & 95.56                                                                 & \textbf{91.11}                                                        & \textbf{93.40}                                                          \\ \bottomrule
\end{tabular}
\vspace{-8mm}
\end{table}

\section{Conclusion}

Data augmentation of time series is challenging due to the high dimensionality of data, the temporal correlation of each dimension, and the difficulty of comparing time series similarity. In this study, we propose a synthetic minority oversampling technique based on the DTW metric, which uses an autoencoder that is unsupervisedly trained to adapt the DTW metric for interpolation. To accommodate the phase shift and amplitude change of the time series, we use DTW as the distance metric. To adapt the interpolation to the DTW distance metric, we use an unsupervised trained siamese network as an encoder so that it can map the time series to the Euclidean deep feature space. After that, we interpolate in the Euclidean deep feature space and use a decoder to form the new synthetic data. We evaluated the effectiveness of DTWSSE on a number of datasets and find that DTWSSE performs better than classical SMOTE in most cases and both the DTW distance metric and the use of interpolation with DTW-adapted autoencoder are essential to obtain better performance.

\section*{Acknowledgements} 
Zixuan Li from Tianjin University and Pengfei Liu from Zhejiang University contributed to this work by providing advice and assistance.

This work is supported by the school-enterprise cooperation project of Yanbian University [2020-15], State Language Commission of China under Grant No. YB135-76 and Doctor Starting Grants of Yanbian University [2020-16].

\bibliographystyle{splncs04} 
\bibliography{refs}

\end{document}